# Recoverability policy editing for closed-loop safety in autonomous driving

Yang Guan[1*], Shengbo Eben Li[1,2*]

[1] School of Vehicle and Mobility, Tsinghua University, Beijing, China

[2] College of AI, Tsinghua University, Beijing, China

* Corresponding authors: yguan@tsinghua.edu.cn, lishbo@tsinghua.edu.cn

## Abstract

Autonomous driving powered by neural network models lacks safety guarantees. We propose a state-level policy editing technique that guarantees closed-loop safety under recoverable disturbances while reusing the model's existing driving capabilities. Closed-loop safety means that the vehicle remains free of collisions at all future times. We introduce the concept of policy recoverability to characterize whether a vehicle can safely return to the original policy's feasible set in finite time. Our method identifies states that are unsafe in the long term under the original policy but recoverable, and selectively edits policy outputs within this region. The vehicle corrects its actions upon entering this region and resumes the original policy upon recovery. Closed-loop simulations on nuPlan urban driving scenarios show a 48% reduction in safety incidents under the original policy while largely preserving its driving performance.

## Introduction

Safety remains a central challenge in realizing autonomous driving. Safety failures directly endanger human lives and cause economic losses. Autonomous driving systems must therefore continually suppress collisions while maintaining high driving performance[1]. The design of these systems is shifting from layers of rules based on expert knowledge towards data-driven neural network models[2,3]. Trained on large amounts of human driving data, these emerging systems have acquired proficient driving capabilities in most scenarios and represent a key route to highly automated driving. Nevertheless, ensuring that neural network models remain safe throughout closed-loop operation is a major challenge. Differences between how the deployment and training environments evolve can push vehicles into out-of-distribution regions with sparse data and insufficient recovery actions[4,5]. As a result, autonomous vehicles may lose the ability to return to safe states, undermining their closed-loop safety[6].

In this work, we use neural network model editing to ensure the closed-loop safety of autonomous driving models under recoverable disturbances, replacing extensive post-processing rules for preventing hazardous incidents. Neural network fine-tuning optimizes a given model on target data while preserving its existing capabilities as much as possible[7]. For a given autonomous driving model, its closed-loop behaviour induces a feasible set[8]. Starting from this set, the original policy can ensure safe driving at every future time. Outside this set, however, the model will eventually drive the vehicle into an unsafe state, even if the vehicle is currently safe. In deployment, the model typically operates within its own feasible set, for example during small steering adjustments and routine interactions. However, controller execution errors and abnormal environmental dynamic mismatch[9] may push the vehicle into states without closed-loop safety guarantees. We require the edited neural policy to have the maximum ability to recover the vehicle from such deviations to feasible states, while leaving the original policy's planning outputs unchanged within its feasible set. This ensures that the autonomous driving model provides closed-loop safety guarantees when disturbances place the state in a recoverable region.

To this end, we propose Recoverability Policy Editing (RPE). The method identifies the largest set of states from which the vehicle can recover to the original policy's feasible set, and edits the neural network model's outputs only for states in this set, driving the vehicle back to the original policy's feasible set and keeping it there (Fig. 1). This approach transforms safety fine-tuning from empirical retraining into policy editing restricted to a state region, allowing closed-loop safety and existing policy capabilities to be maintained under recoverable environmental disturbances. Our technique has three key features:

1. Recoverable set identification: we assess the closed-loop safety of each state under the original policy. After determining its feasible set, we solve for the set of states that are unsafe in the long term under the original policy but can safely return to its feasible set through policy editing. Compared with conventional fine-tuning methods, our approach precisely characterizes the boundary of the largest set of states over which existing capabilities can be reused, enabling policy fine-tuning targeted at this set.

2. Policy editing for recovery: our method ensures that the edited policy can drive the autonomous vehicle from any state in the recoverable set back to the original policy's feasible set in the shortest possible time, and keep it there. Fine-grained policy editing guarantees closed-loop safety while making full use of the original policy's capabilities, provided that each environmental disturbance leaves the state within the union of the feasible and recoverable sets.

3. Preservation by construction: RPE keeps the original policy fixed and learns a gated residual branch. When the gate is closed, the edited policy reproduces the original outputs exactly. By restricting residual corrections to recoverable states outside the original policy's feasible set, the formulation preserves the original policy's behaviour within that set. The formulation accommodates both learned policies and optimization-based trajectory planners.

Existing methods for ensuring closed-loop safety in autonomous driving fall into two categories: safety post-processing and safe policy learning. The former primarily checks candidate trajectories at runtime and overrides the original policy's outputs when they are deemed hazardous. Formalized road rules provide safety specifications for online checks[10], and online trajectory verification checks whether candidate trajectories satisfy these specifications at runtime[11]. Runtime assurance[12], shielding[13], predictive safety filtering[8] and shielding mechanisms for interactive planning[14] assess and correct potentially hazardous actions before execution. These methods can prevent hazardous actions from being executed under given safety specifications, but do not focus on whether the driving state after intervention is recoverable with respect to the original policy. Safe policy learning methods often introduce closed-loop safety conditions during training, such as reachability analysis[15], control barrier functions[16], Lyapunov functions or invariant sets[17], and neural certificates[18], to prevent the system from entering unsafe states or ensure that the vehicle remains within a safe invariant set. These conditions are further incorporated as constraints into safe reinforcement learning to train neural policies[19,20,21,22,23], avoiding the substantial online computational burden associated with complex traffic safety constraints. However, these methods typically require learning a control policy from scratch. For safety repair of an existing driving model, how these safety conditions connect to the original policy's range of capabilities remains unclear.

Autonomous driving commonly relies on foundation models trained on vast amounts of driving data. In this context, the challenge is to repair unsafe behaviour while preserving the model's existing driving capabilities. Existing methods typically formulate preference or reward objectives that suppress collisions and fine-tune a given model using direct preference optimization[24,25] or reinforcement learning[26,27,28,29]. To mitigate the loss of the given model's capabilities, these methods commonly introduce pretraining losses[30], parameter-efficient modules[31] or regularization of deviations from the initial model[32]. However, these methods are simultaneously driven by safety and capability-preservation objectives across the entire state distribution. They neither provide theoretical closed-loop safety guarantees nor eliminate catastrophic forgetting. Previous studies[33,34,35,36] fine-tune models using safety-critical scenarios in which a given model causes collisions or violations. Such hazardous scenarios may still contain states that the original policy can handle safely, as well as states from which a safe return to its range of capabilities is impossible. Scenario-level selection cannot adequately distinguish the editing needs of these states. Policy updates under a particular distribution may also affect outputs in other states[37]. We therefore address a central question: to improve a given model's closed-loop safety, in which states should the original policy's outputs be preserved, in which states should they be optimized, and how should this optimization be performed? Our method precisely edits only states that are infeasible under the original policy but remain recoverable to its feasible set, enabling the fine-tuned policy to return quickly to that set.

Safely recovering the driving state requires resolving the coupling between the policy-induced recoverable set and the policy itself: the policy determines the extent of the recoverable set, yet must itself be edited within that set. Existing work often directly defines the driving states that require intervention under the original policy and corrects candidate actions through reachability-based safety layers, action projection or online verification, enabling the system to avoid collisions and return to safe states[38,39,40,41]. Such methods can block locally unsafe actions, but they typically rely on external safety certificates or reachability analyses to determine when to intervene. These criteria do not necessarily characterize the maximal recoverable set induced by the original policy. Specifically, methods based on barrier functions typically correct potentially hazardous actions in advance[42,43,44]. Their admissible sets are constrained by the form of the certificate function and constraint thresholds, often leading to a conservative trade-off between provable safety and minimal intervention[16]. Another class of methods switches to a recovery policy or a safety layer under hazardous conditions[21,12,11,22]. However, these methods generally focus on satisfying immediate safety constraints without considering the original policy's closed-loop safety at the current state, which may cause them to miss the best time to intervene. Our method introduces an iterative procedure for jointly solving for the edited policy and the recoverable set, ensuring the maximal extension of the original policy's feasible set while obtaining an edited policy that achieves the fastest recovery.

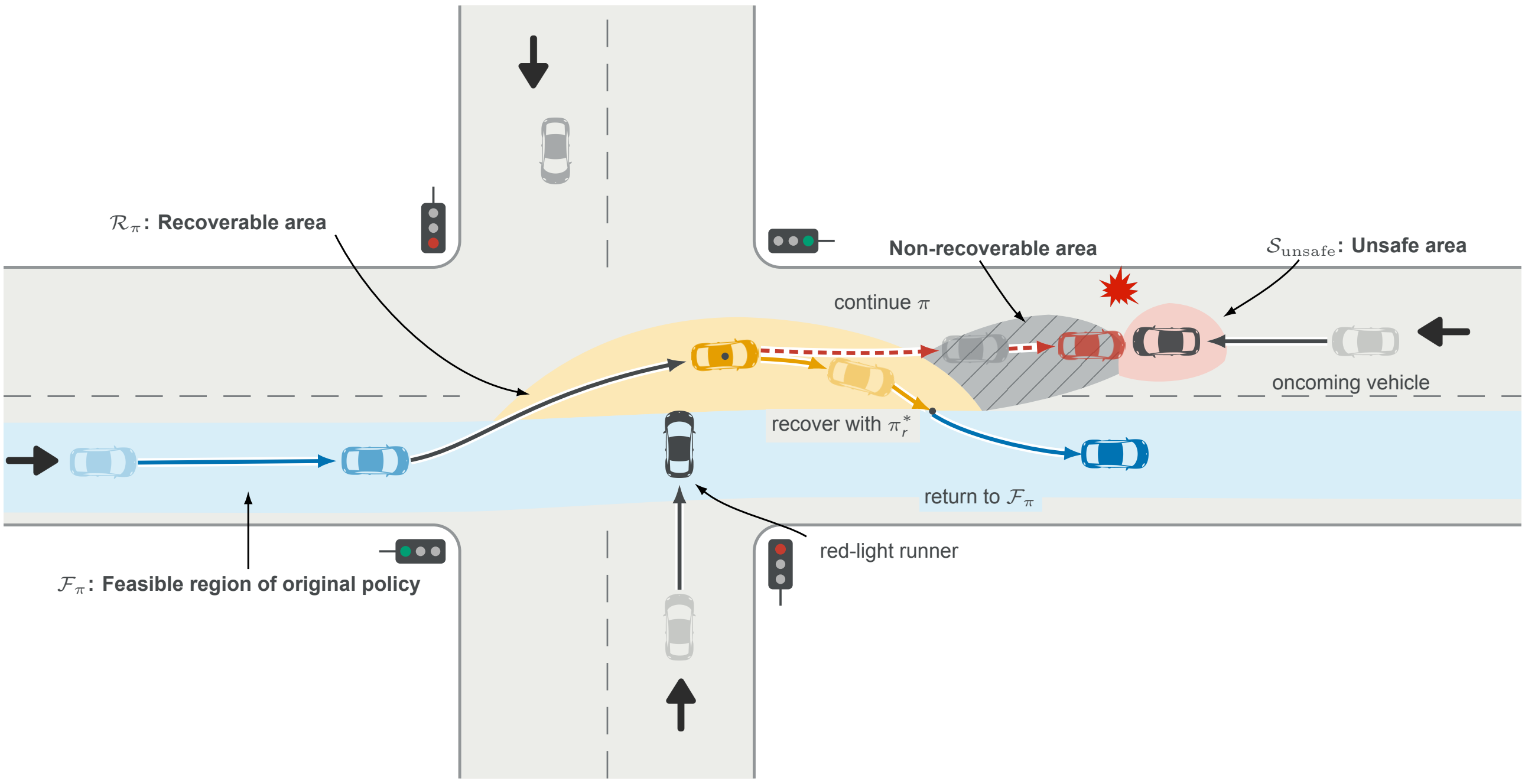

**Figure 1. The concept of a policy's recoverable set.** The feasible set of a given neural policy (blue region) guarantees the vehicle's closed-loop safety. However, disturbances whose evolution differs from the training environment, such as an unexpected vehicle running a red light, can force the vehicle outside this set. Without policy intervention, the vehicle will eventually enter an unsafe state (red region). While preserving the original policy's capabilities, our method first identifies the set of states that are recoverable to the original policy's feasible set (yellow region) and those that are not (grey region). It then incrementally edits the original policy's outputs in recoverable states, guiding the vehicle back into the original policy's range of capabilities and thereby ensuring closed-loop safety.

To provide closed-loop safety guarantees for a given policy, our central idea is to define the original policy's recoverable set, precisely edit neural network outputs at the state level within this set, and ensure that outputs remain unchanged in states within the feasible set (Fig. 2a-2c). We define feasibility and recoverability under deterministic dynamics $s_{t+1} = f(s_t, a_t)$, with $f_\pi(s) = f(s, \pi(s))$ under the original policy. For an original policy $\pi$, its recoverable set $\mathcal{R}_\pi$ consists of states outside its feasible set for which there exists an edited policy $\pi_r$ that can safely return the vehicle to the original policy's feasible set in finite time:

$$\mathcal{R}_\pi = \{s \in \mathcal{S} \setminus \mathcal{F}_\pi \mid \exists \pi_r \in \Pi,\ \text{s.t. using } \pi_r, s \text{ safely leads to } \mathcal{F}_\pi \text{ in finite time}\} \tag{1}$$

Here, $\mathcal{F}_\pi$ is the feasible set of the original policy $\pi$, defined as $\mathcal{F}_\pi = \{s \mid s_t \in \mathcal{S}_{\text{safe}}, a_t = \pi(s_t), t = 0, 1, \ldots, \infty, s_0 = s\}$. $\mathcal{S}_{\text{safe}}$ denotes the set of safe states. $\Pi$ is the set of candidate edited policies, whose gated residual structure is introduced below. We call a disturbance recoverable with respect to policy $\pi$ if the resulting state lies in $\mathcal{F}_\pi \cup \mathcal{R}_\pi$.

Our method first applies fixed-point iteration of the Bellman operator $\mathcal{T}_\pi$ to progressively remove states from the safe state set that are unsafe in the long term under policy $\pi$. This yields an estimate of the feasible set, $\mathcal{F}_\pi = \{s \in \mathcal{S}_{\text{safe}} : F_\pi(s) = 1\}$, where $F_\pi(s)$ is the fixed point of $\mathcal{T}_\pi$ and indicates the state's closed-loop safety (Fig. 2d).

$$(\mathcal{T}_\pi F)(s) = \begin{cases} 0, & s \in \mathcal{S} \setminus \mathcal{S}_{\text{safe}}, \\ F(f_\pi(s)), & s \in \mathcal{S}_{\text{safe}} \end{cases}, F^0(s) = \mathbf{1}\{s \in \mathcal{S}_{\text{safe}}\} \tag{2}$$

We then construct the edited policy using a residual branch and a gate, $\pi_r(s) = \pi(s) + g(s)\delta(s)$. Here, $\pi$ is the original policy before editing, whose parameters remain frozen during training, and $\delta$ is a trainable residual branch that outputs corrections to the original policy's actions. When the gate is closed, the edited policy preserves the original policy's outputs. When the gate is open, residual corrections enable safe recovery until the state returns to the original policy's feasible set, at which point the gate closes and the original policy's outputs are used again. The gating function $g(s)$ controls the scope of policy editing: it equals 1 when $s \in \mathcal{R}_\pi$ and 0 otherwise. This leaves the original policy's outputs unchanged within the feasible set and enables policy editing through the residual branch within the maximal recoverable set. We use an iterative procedure to resolve the coupling between the recoverable set and the edited policy, first defining the recoverability indicator function of the edited policy $\pi_r$ with respect to the original policy $\pi$:

$$G_\pi^{\pi_r}(s) = \beta^{\tau_{\mathcal{F}_\pi}} \mathbf{1}\{\tau_{\mathcal{F}_\pi} < \tau_{\mathcal{S}_{\text{unsafe}}}\} \tag{3}$$

Here, $\tau_{\mathcal{F}_\pi}$ is the first time that the trajectory induced by the edited policy $\pi_r$ from $s$ enters the feasible set, $\tau_{\mathcal{S}_{\text{unsafe}}}$ is its first entry time into the unsafe set, and $\beta \in (0, 1)$ is a discount factor. A larger $G_\pi^{\pi_r}(s)$ indicates that, starting from state $s$, the edited policy returns to the feasible set more quickly. When $s \in \mathcal{F}_\pi$, $G_\pi^{\pi_r}(s) = 1$. When $s \in \mathcal{S}_{\text{unsafe}}$, $G_\pi^{\pi_r}(s) = 0$. The recoverable set of $\pi_r$ associated with this indicator function can accordingly be expressed as $\{s \in \mathcal{S} \setminus \mathcal{F}_\pi : G_\pi^{\pi_r}(s) > 0\} \subseteq \mathcal{R}_\pi$.

To jointly obtain the optimal recoverability indicator function, $G_\pi^*(s) = \sup_{\pi_r \in \Pi} G_\pi^{\pi_r}(s)$, and the corresponding optimal edited policy $\pi_r^*$, we design a recovery Bellman operator:

$$(\mathcal{T}_{\pi,\text{rec}} G)(s) = \begin{cases} 1, & s \in \mathcal{F}_\pi, \\ 0, & s \in \mathcal{S}_{\text{unsafe}}, \\ \beta \max_{a \in \mathcal{A}} G(f(s,a)), & \text{otherwise} \end{cases} \tag{4}$$

This operator is a contraction mapping with the unique fixed point $G_\pi^*$. The fixed point is obtained by iteratively optimizing the recoverability indicator function and the edited policy. Specifically, given the edited policy $\pi_r^k$ at iteration $k$, we replace the maximization term in the recovery Bellman operator with this policy and iteratively solve for its recoverability indicator function $G_\pi^k$. We then optimize the edited policy to maximize the indicator values over $s \in \mathcal{S} \setminus (\mathcal{F}_\pi \cup \mathcal{S}_{\text{unsafe}})$, that is, $\max_{\pi_r} G_\pi^k(f(s, \pi_r(s)))$, obtaining $\pi_r^{k+1}$. The process continues until convergence, namely $G_\pi^k(s) = G_\pi^{k+1}(s), \forall s \in \mathcal{S}$. At this point, $G_\pi^k$ is the optimal recoverability indicator function $G_\pi^*$, and $\pi_r^k$ is the edited policy $\pi_r^*$ that achieves the fastest recovery (Fig. 2e).

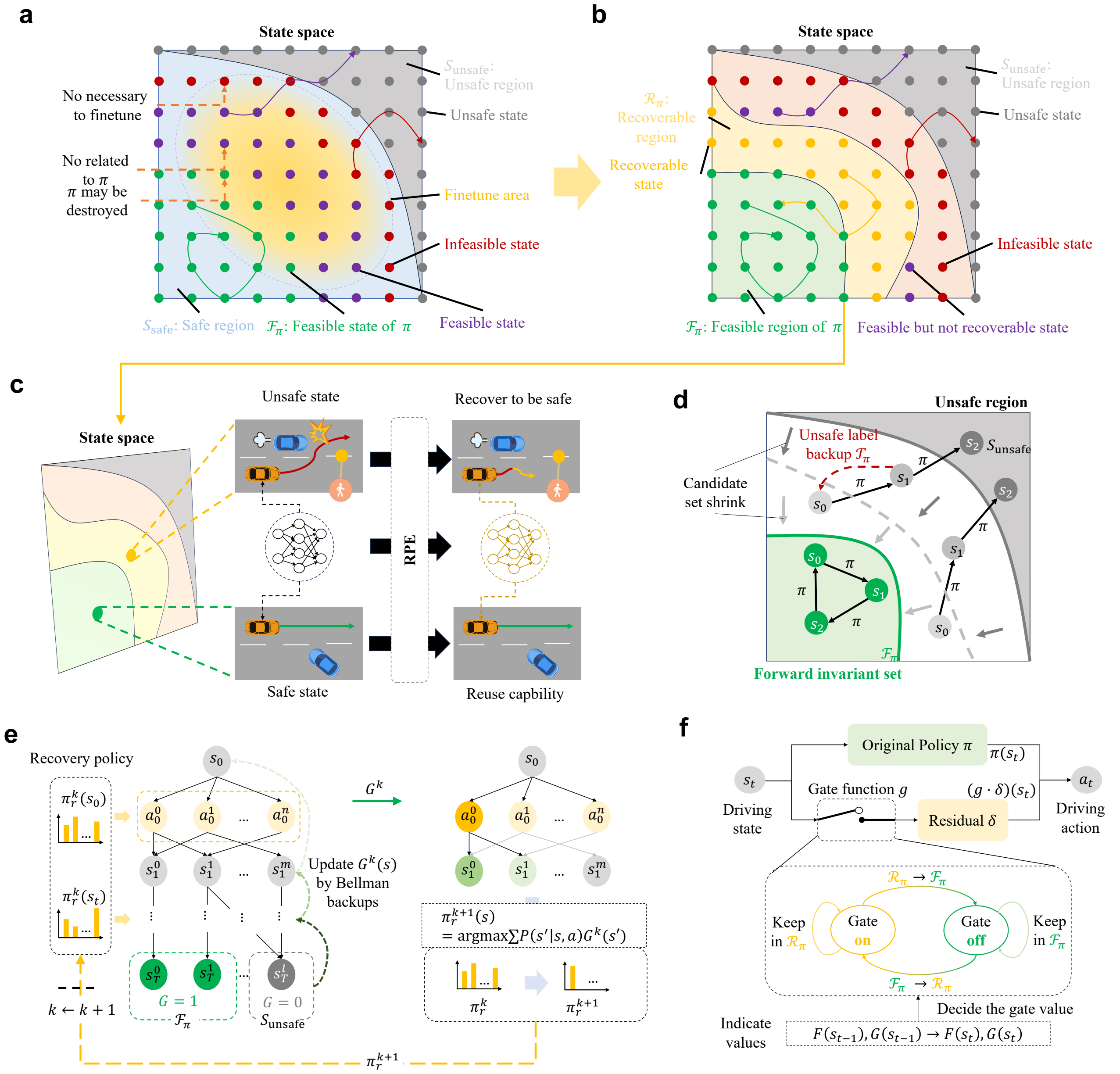

**Figure 2. State-level policy editing within the recoverable set. a**, Fine-tuning that does not distinguish state recoverability may affect states that need no modification or cannot be recovered, and may undermine the original policy's existing capabilities. **b**, Identifying the original policy's feasible set $\mathcal{F}_\pi$ and recoverable set $\mathcal{R}_\pi$ determines the scope of policy editing. **c**, State-level policy editing repairs behaviour in recoverable states while reusing the original policy's capabilities within the feasible set. **d**, Estimation of the original policy's feasible set. Unsafe labels are propagated backwards along state transitions under the original policy, iteratively shrinking the candidate set until it converges to a forward-invariant set of states. **e**, Solution of the edited policy. The recoverability indicator function and edited policy are updated alternately. The former uses a Bellman operator to bootstrap state values, with larger values indicating faster recovery to the original policy's feasible set. The latter updates the policy to maximize these values. This process continues until convergence. **f**, Online decision-making with the edited policy. The feasibility indicator $F$ and recoverability indicator $G$ determine whether the residual gate is open or closed. When the state leaves the feasible set and remains recoverable, the gate opens, and the edited policy uses the residual branch to correct the original policy's outputs and perform safe recovery. When the state returns to the feasible set, the original policy's outputs are restored.

Finally, we apply the optimized edited policy $\pi_r^*$ to online decision-making (Fig. 2f). Specifically, when the vehicle state lies within the feasible set $\mathcal{F}_\pi$, decisions are made using the original policy. When a disturbance inconsistent with the training environment pushes the vehicle outside the feasible set but leaves its state in the recoverable set, the gate opens and the residual branch corrects the original policy's outputs to perform safe recovery. The gate remains open until the state re-enters the original policy's feasible set. This process is controlled by the following switching conditions for the gating function $g(s)$:

$$g(s) = \begin{cases} 1 \to 0, & F_\pi(s) = 1, \\ 0 \to 1, & F_\pi(s) = 0 \text{ and } G_\pi^*(s) > 0 \end{cases} \tag{5}$$

Here, $F_\pi(s) = 0$ and $G_\pi^*(s) > 0$ are equivalent to $s \in \mathcal{R}_\pi$. Thus, the gate opens when the state enters the maximal recoverable set and closes when it returns to the feasible set. If neither switching condition is satisfied, the gate retains its current state. The edited policy achieves the fastest recovery while the gate is open and reuses the original policy after returning to the feasible set, thereby enabling safe recovery while preserving the original policy's capabilities.

In the implementation, we use neural networks to approximate the state feasibility and recoverability indicator functions, denoted by $F(s)$ and $G(s)$, respectively. The recoverability indicator function is optimized using a TD-Lambda target that combines multistep returns, and the residual branch is updated using proximal policy optimization. During online operation, the gate determines when to intervene and when to withdraw based on the network estimates and specified thresholds. Network training and gating settings are described in Methods.

## Results

We examine RPE's closed-loop safety and preservation of the original policy's capabilities through three experiments. First, we record changes in the state regions and policy during alternating iterations of the recoverability indicator function and the edited policy in a two-dimensional driving scenario. Second, we perform closed-loop simulations in hazardous scenarios drawn from real-world driving logs to analyse the effects of local policy editing. Finally, we statistically assess the edited policy's closed-loop safety and preservation of the original policy's capabilities through closed-loop simulations on 1,708 nuPlan scenarios.

### Joint learning of recoverability and policy

We first study how the recoverable set and edited policy evolve during training. The driving scenario is shown in Fig. 3a. The ego vehicle travels on a highway with an on-ramp and a construction zone on the main road. Its longitudinal speed is constant, and only its lateral velocity is controllable. The original policy regulates lateral velocity through negative feedback on lateral displacement only near the main road's centreline, enabling lane-centred driving. Elsewhere, it continually moves away from the centreline. For comparison, we also solve for the various safety-related regions in this driving scenario, all of which can be represented as sets of states in a two-dimensional plane.

**Training of the original policy's feasible set.** Figure 3b shows the training of the feasibility indicator function associated with the original policy's feasible set $\mathcal{F}_\pi$. The feasible set shrinks monotonically from the full state space to a region near the lane centreline of the main road. The safe set $\mathcal{S}_{\text{safe}}$ and maximal feasible set $\mathcal{F}^*$ for this scenario are also shown. The maximal feasible set indicates states for which a policy exists that can maintain the vehicle's closed-loop safety. The results show that the given policy's feasible set is a subset of the safe set, and that states outside it will eventually lead the vehicle into unsafe states. Meanwhile, closed-loop safe policies exist for some states outside the given policy's feasible set, allowing driving safety to be improved through policy editing. However, these set-based methods cannot distinguish which states can return to the original policy's feasible set or provide the corresponding recovery behaviours, making it difficult to preserve the original policy's capabilities.

**Training of the recoverable set and the edited policy.** Based on the original policy's feasible set, we use an iterative procedure to alternately train the recoverability indicator network and the residual branch of the edited policy. Figure 3c shows how the mean value of the state recoverability indicator function changes during training. This value increases as training progresses until convergence to the optimal recoverability indicator function. Figure 3d shows the changes in the recoverability indicator function, recoverable set and edited policy during training.

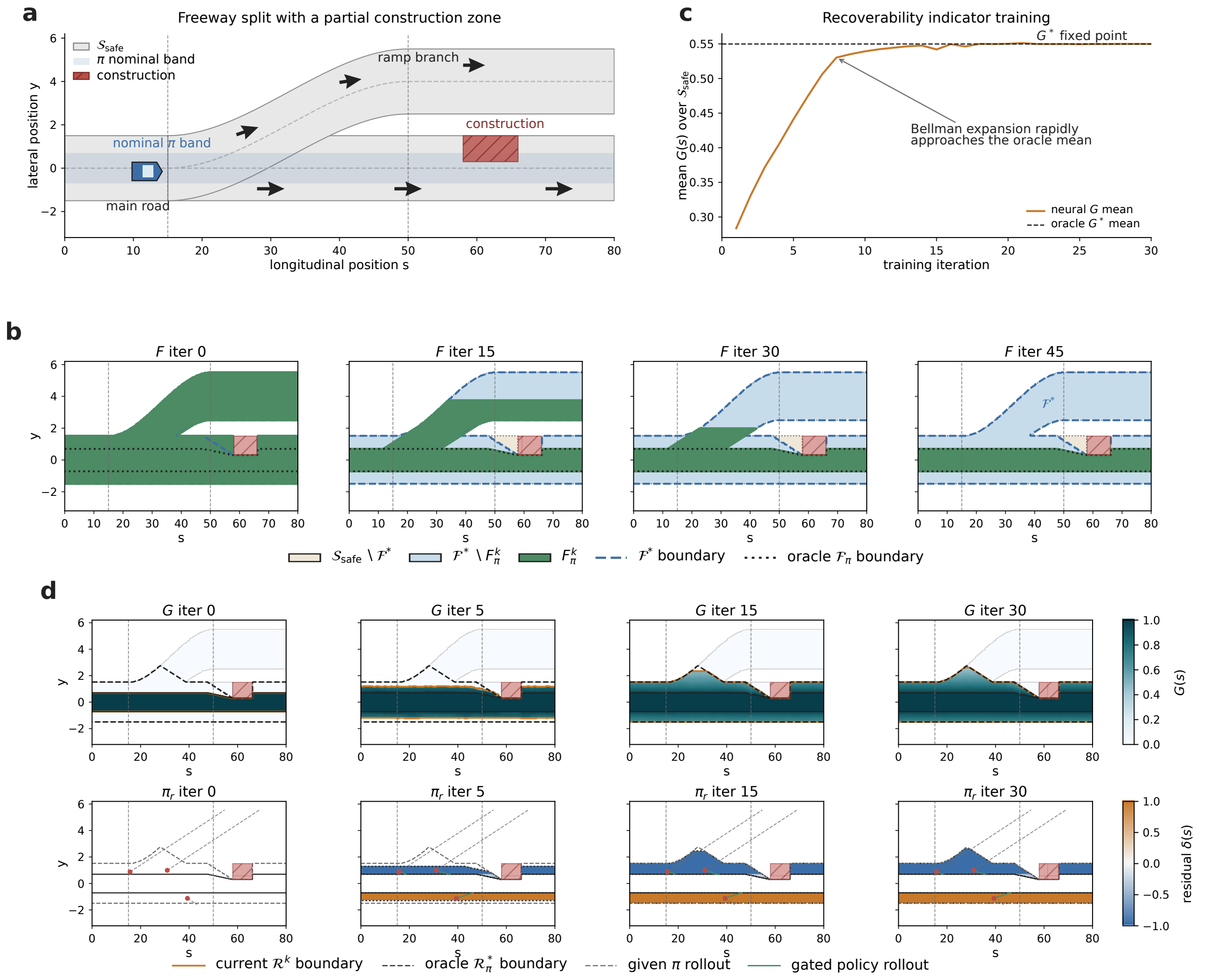

**Figure 3. Training of the maximal recoverable set and the edited policy. a**, A two-dimensional driving scenario with an on-ramp and a construction zone, together with the original policy. **b**, Training of the original policy's feasible set $\mathcal{F}_\pi$. **c**, Training of the recoverability indicator function $G_\pi^*$. **d**, Evolution of the recoverable-set boundary and the edited policy during training.

As training progresses, the recoverable set gradually expands outwards from the original policy's feasible set until it reaches the maximal recoverable set. The indicator network outputs smaller values for states farther out, indicating that more steps are required for recovery. Meanwhile, some states on the on-ramp lie within the maximal feasible set but are not recoverable to the original policy's feasible set. Our method targets policy optimization at states within the recoverable set, avoiding interference from ineffective policy optimization in other regions. Within this set, the policy changes guide the vehicle quickly back to the original policy's feasible set.

## Policy editing in a hazardous case

We further investigate whether local policy editing can repair the original policy's driving behaviour in a hazardous scenario while allowing the original policy to be reused after intervention. The scenario is drawn from real-world driving logs in the nuPlan dataset and was not used to train the edited policy. As shown in Fig. 4a, the ego vehicle approaches a red traffic light, and a vehicle from the side subsequently enters its path. This incursion creates a potential conflict between the vehicles' motions, requiring the ego vehicle to adjust its trajectory promptly. We compare the driving performance of the original and edited policies in closed-loop simulations with identical traffic participant replay conditions.

**Driving performance of the original policy.** The red dashed lines in Fig. 4b show the estimated feasibility indicator $F$, estimated recoverability indicator $G$ and gate $g$ over time during driving with the original policy. The estimated feasibility indicator is close to 1 early in the scenario but decreases continuously as the side vehicle approaches. At $t = 9.0\,\mathrm{s}$, it falls below the threshold of 0.3 used to estimate the feasible set and remains below it thereafter. This indicates a decline in the network's estimate of the original policy's future safety. The vehicle eventually enters an unsafe state at $t = 14.1\,\mathrm{s}$. Meanwhile, the recoverability indicator remains low, indicating that recovery to the original policy's feasible set is difficult.

The second row of Fig. 4c shows the original policy's closed-loop behaviour at key moments. The ego vehicle initially approaches the red-light intersection at $7.7\,\mathrm{m/s}$, while the side vehicle moves towards the main road and then stops to let the ego vehicle pass. However, the ego vehicle continues to decelerate in response to the red light. At $11.3\,\mathrm{s}$, the side vehicle starts moving. The original policy decelerates more strongly but still fails to avoid the conflict with the side vehicle, eventually causing a collision.

**Selective intervention by the residual branch.** The blue solid lines in Fig. 4b show the estimates $F$ and $G$ and the gate during driving with the edited policy. During the initial period of $1.0$-$9.0\,\mathrm{s}$, the gate remains closed and the vehicle directly uses the original policy. Subsequently, when the state is classified as entering the recoverable set ($F < 0.3$ and $G \geq 0.1$), the controller opens the gate, and the edited policy corrects the original policy's output trajectory through the residual branch. When the state is classified as having returned to the original policy's feasible set ($F > 0.9$), the gate closes and the original policy is used again. During this process, the residual branch intervenes only in three windows: $9.0$-$9.9\,\mathrm{s}$, $11.4$-$12.3\,\mathrm{s}$ and $12.9$-$13.2\,\mathrm{s}$. Both $F$ and $G$ increase substantially following intervention and differ markedly from the original policy's estimates, indicating that the state gradually returns to the original policy's feasible set during intervention.

The first row of Fig. 4c shows the closed-loop simulation results for the edited policy. During $6.0$-$9.0\,\mathrm{s}$, the ego vehicle's speed continues to decrease and the interaction risk increases, prompting the residual branch to intervene and the ego vehicle to accelerate. At $9.9\,\mathrm{s}$, the safety risk decreases and the original policy resumes. However, at $11.3\,\mathrm{s}$, the side vehicle starts moving, and state feasibility and recoverability decrease sharply. The residual branch intervenes again, causing the ego vehicle to accelerate until $12.3\,\mathrm{s}$, after which the original policy resumes. At $12.9\,\mathrm{s}$, increased acceleration of the side vehicle reduces state feasibility again, while recoverability decreases only partially. The residual branch intervenes briefly once more until $13.2\,\mathrm{s}$. The gate then remains closed, and the vehicle continues using the original policy until the end of the scenario. These interventions increase the ego vehicle's speed from $7.3\,\mathrm{m/s}$ to $8.2\,\mathrm{m/s}$, which is $2.3\,\mathrm{m/s}$ faster than the original policy at the time of collision, enabling successful passage through the interaction area.

**The role of local editing in safety repair.** Figure 4d further shows the trajectory modifications made during the three interventions: at the same state, the residual branch applies longitudinal and lateral increments to the original policy's output trajectory. The increment control points are at $0.5/1.0/1.5/2.0\,\mathrm{s}$. After interpolation, the increments decay to zero before $3.0\,\mathrm{s}$. Across the three interventions, the increment control points shift leftwards laterally, increasing the clearance between the ego and side vehicles. Longitudinally, the changes mainly affect the control points at $1.0\,\mathrm{s}$ and $1.5\,\mathrm{s}$, increasing the speed at the $1\,\mathrm{s}$ trajectory point, which is the preview target of the downstream trajectory-tracking controller. This shows that the edited policy modifies the original policy's trajectories through the residual branch, thereby affecting the vehicle's closed-loop driving behaviour.

To test the effectiveness of these brief interventions, we remove the residual in W1, W2 or W3 individually, hold the timing of the other gating windows fixed, and rerun the closed-loop simulations. Figure 4e shows that removing either W1 or W2 causes a collision. Removing W3 still allows the vehicle to pass through the segment, but reduces the minimum clearance from $0.531\,\mathrm{m}$ with all interventions to $0.447\,\mathrm{m}$. Thus, under this fixed-window comparison, the first two interventions play a key role in avoiding collision, whereas the third increases the safety margin. These results show that brief, local policy editing can effectively repair the original policy's collision behaviour.

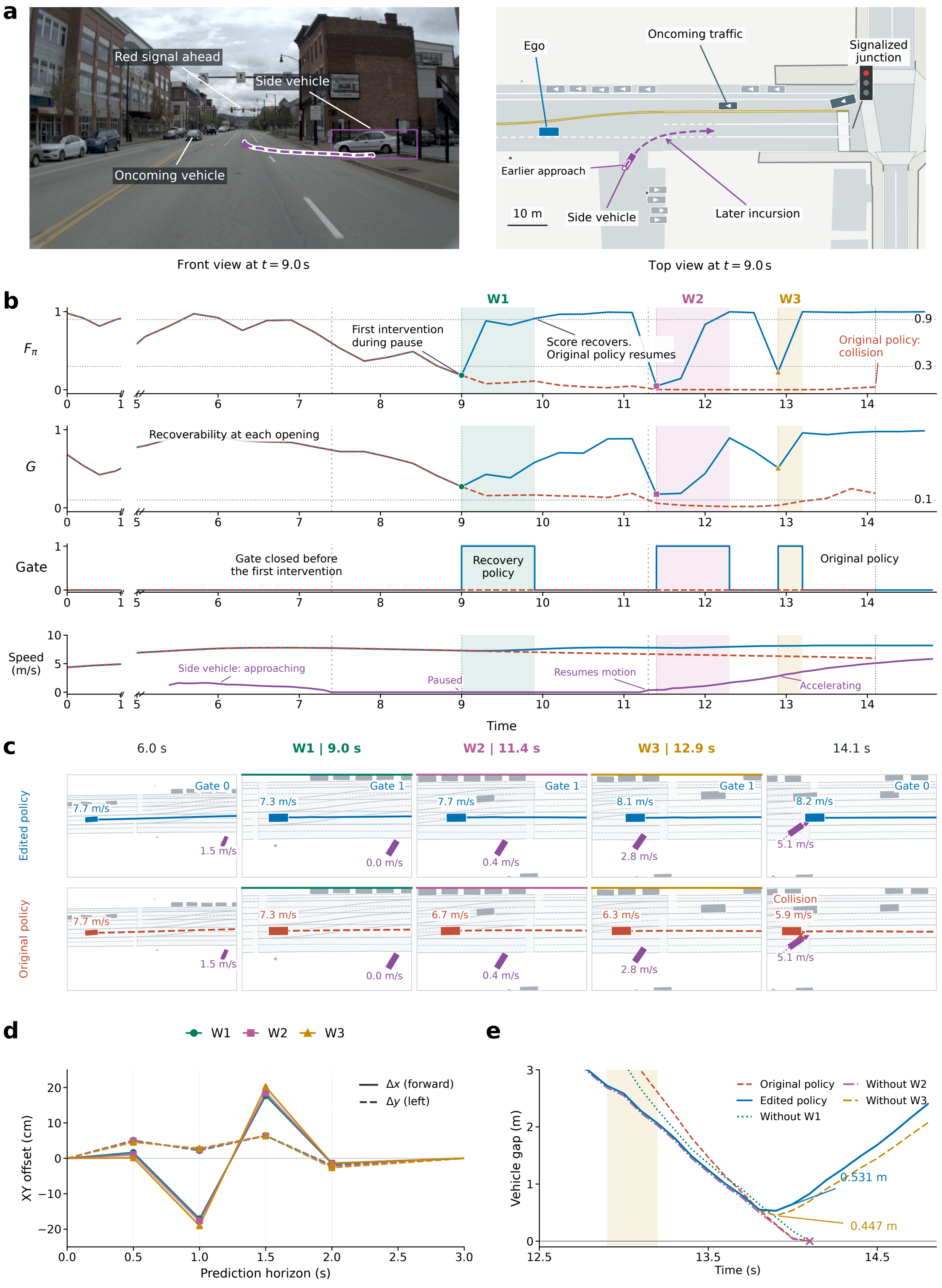

**Figure 4. Local policy editing in a hazardous scenario. a**, A hazardous interaction from real-world nuPlan logs. The ego vehicle approaches a red light as a side vehicle enters its path. The purple dashed line shows the side vehicle's recorded future trajectory. **b**, Neural network estimates of $F_\pi$ and $G^*_\pi$, together with gate $g$ and vehicle speed over time. W1-W3 denote three brief interventions by the residual branch. **c**, Closed-loop simulation snapshots. The original policy collides, whereas the edited policy passes safely and resumes the original policy after each intervention. **d**, Longitudinal and lateral trajectory increments relative to the original policy at the same state. Colours denote intervention windows. Solid and dashed lines denote longitudinal and lateral increments, respectively. **e**, Inter-vehicle distances when one intervention is removed while the other gating windows remain fixed. Removing W1 or W2 causes a collision. Removing W3 reduces the minimum clearance without causing a collision.

## Safety repair and capability preservation

Editing the original policy requires assessing both its ability to repair hazardous scenarios and whether its original driving performance is preserved. We further test the safety repair capability of the gated edited policy trained by RPE, together with its preservation of the original policy's performance, on 1,708 nuPlan scenarios. As shown in Fig. 5a, the test set includes various types of long-tail and hazardous scenarios. The original policy exhibits unsafe behaviour, involving collisions or departures from the drivable area, in 244 scenarios and remains safe in the remaining 1464. For comparison, we use conventional RL to update all network parameters of the original policy to produce safer actions. This method uses the same training data as RPE, comprising scenarios in which the original policy drives unsafely.

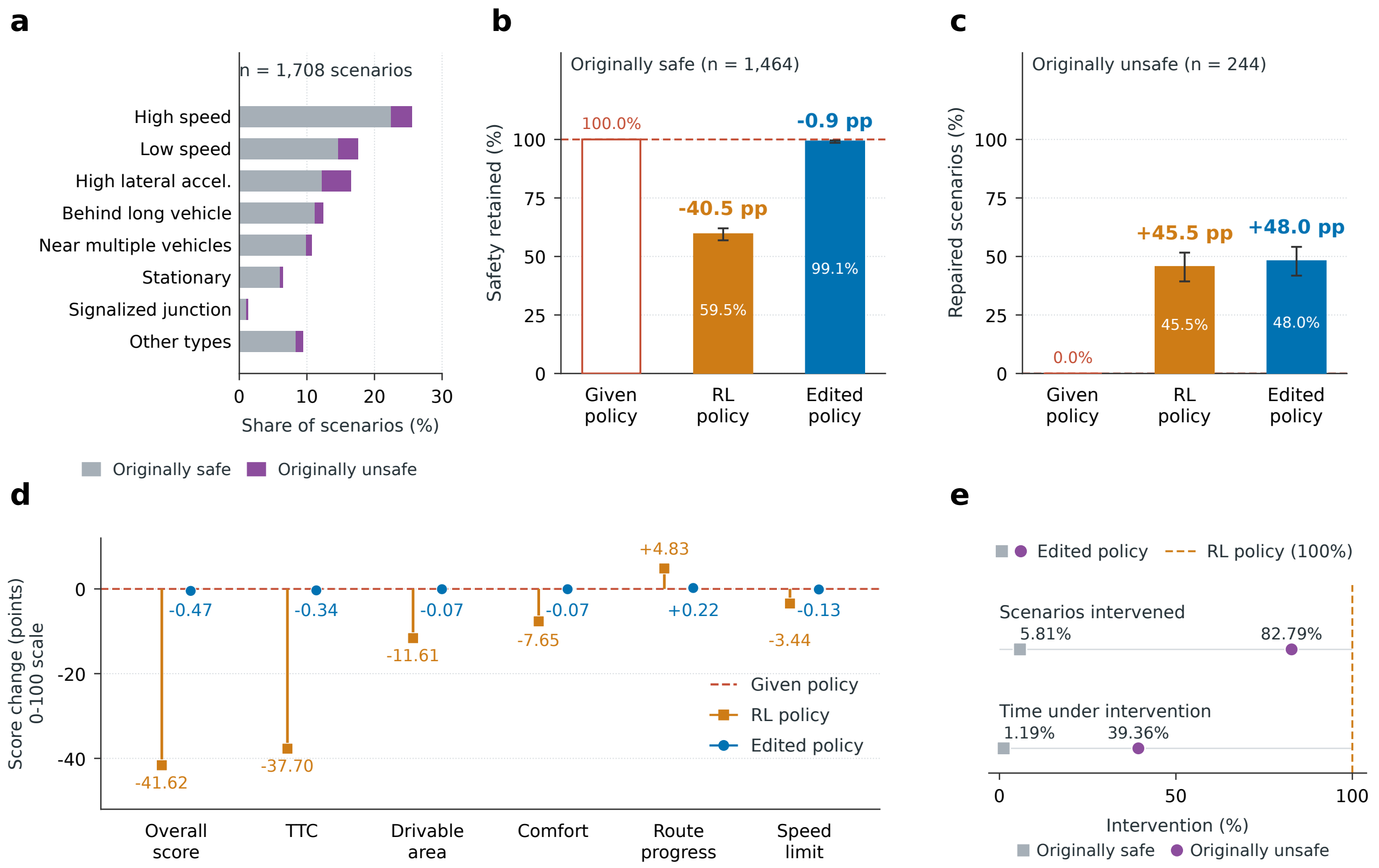

**Figure 5. Safety repair and capability preservation through selective policy editing. a**, Distribution of test scenario types. Grey and purple denote originally safe and originally unsafe scenarios, respectively. **b**, Safety retention rate on originally safe scenarios. **c**, Repair rate on originally unsafe scenarios. In b and c, smaller labels show absolute percentages and bold labels show percentage-point changes relative to the original policy. Error bars are 95% percentile confidence intervals from 10,000 paired bootstrap resamples. **d**, Mean changes in six nuPlan metrics relative to the original policy on originally safe scenarios, ranging from -100 to 100. Orange squares and blue circles denote the RL policy and edited policy, respectively. The red dashed line denotes the original policy's zero-change reference. **e**, Fraction of scenarios with at least one intervention and fraction of time spent intervening. Grey squares and purple circles denote the edited policy's results on the originally safe and originally unsafe groups, respectively. The orange dashed line denotes the always-active RL policy (100%). The intervention time fraction is calculated as total gate-open time divided by total observation time.

Figures 5b and 5c show the safety retention rate on originally safe scenarios and the repair rate on originally unsafe scenarios, respectively. The edited policy repairs 48.0% (117/244) of originally unsafe scenarios and remains safe in 99.1% (1,451/1,464) of originally safe scenarios. This indicates that RPE can substantially repair safety in the original policy's recoverable states while preserving safety in states within its feasible set. By contrast, although the RL policy is optimized specifically for safety metrics, its safety repair rate does not improve over that of the edited policy and is even lower (45.5%). More importantly, its safety retention rate on originally safe scenarios is only 59.5% (871/1,464), 39.6 percentage points below that of the edited policy. This indicates that, despite training on the same unsafe scenario data, the RL policy's parameter changes have severely damaged the original policy's existing capabilities.

Figure 5d shows the changes in nuPlan driving metrics for RPE relative to the original policy on the safe scenario set, including the overall driving score and component scores for risk, departures from the drivable area, comfort, efficiency and compliance. Each score ranges from 0 to 100, with higher values being better. For the edited policy, changes in the mean overall and component scores are all within 0.5 points. These results suggest that RPE preserves both safety and broader driving performance within the original policy's feasible set. By contrast, the RL policy markedly changes the original policy's capability profile: although efficiency improves, scores for risk, comfort and compliance decline substantially.

Figure 5e shows the fractions of scenarios and time in which the edited policy's residual branch intervenes in the original policy, for originally safe and originally unsafe scenarios. The RL policy is active throughout closed-loop operation and is shown as a 100% reference. The edited policy's residual gate opens when $F < 0.3$ and $G \geq 0.1$ and remains active until $F > 0.9$. In the originally safe group, interventions occur in 5.81% of scenarios and account for 1.19% of total observation time, indicating infrequent and brief intervention by the edited policy. In the originally unsafe group, the corresponding fractions are 82.79% and 39.36%. These results show that, based on estimates of state safety and recoverability, the edited policy intervenes less often in originally safe scenarios to preserve the original policy's capabilities, while intervening more frequently in originally unsafe scenarios to repair its safety.

## Methods

### Problem formulation

Our work starts with a given autonomous driving policy $\pi$ that already provides strong driving performance in most driving states $s \in \mathcal{S}$ and typically operates within its own feasible set $\mathcal{F}_\pi$, where it can remain safe in the long term. However, differences between how the deployment and training environments evolve can cause the vehicle to leave this set. Our method aims to preserve and reuse the given policy's inherent capabilities to the greatest extent possible and optimize safety on this basis, rather than train a driving policy anew. It requires no complex driving performance metrics. Only when a disturbance pushes the state outside the policy's feasible set does it edit the policy's outputs to return the vehicle quickly to the given policy's range of capabilities, ensuring closed-loop safety under any recoverable disturbance.

We describe autonomous driving as a Markov decision process with deterministic transitions $s_{t+1} = f(s_t, a_t)$. The objective of our method can be formulated mathematically as follows:

$$\begin{aligned} \text{maximize} \quad & J(\pi_r) = \mathbb{E}_{s_0 \sim \mathcal{S}} \left[ \sum_{t=0}^{\infty} \beta^t \mathbf{1}\{s_t \in \mathcal{F}_\pi\} \right] \\ \text{subject to} \quad & \pi_r(s) = \pi(s), \quad \forall s \in \mathcal{F}_\pi \\ & a_t = \pi_r(s_t), \quad \forall t = 0, 1, \ldots, \infty \end{aligned} \tag{6}$$

Here, $\mathbb{E}$ denotes expectation over the sampled initial state, $\mathcal{S}$ is the full state space, and $\mathcal{F}_\pi$ is the feasible set of the given policy $\pi$. The objective in equation (6) is to find an edited policy $\pi_r$ that minimizes the time required for the vehicle to return to the original policy's feasible set from any state. A shorter return time incurs less discounting and yields a higher objective value. The constraints ensure that the edited policy leaves the original policy's outputs unchanged within its feasible set. For states already in $\mathcal{F}_\pi$, the constraints make the objective constant, so no optimization is needed. Conversely, the state space contains states that cannot be improved, including the unsafe state set $\mathcal{S}_{\text{unsafe}}$, the set of safe but infeasible states, and the set of safe and feasible but unrecoverable states. The objective is zero for all states in these sets because they cannot be safely recovered to the original policy's feasible set. Optimizing them would instead introduce update noise and hinder policy optimization in the states that truly matter.

To this end, we define the recoverable set $\mathcal{R}_\pi$ as in equation (1). This definition allows the original problem in equation (6) to be equivalently reformulated as policy optimization over the maximal recoverable set:

$$\begin{aligned} \text{maximize} \quad & J(\pi_r) = \mathbb{E}_{s_0 \sim \mathcal{R}_\pi} \left[ \sum_{t=0}^{\infty} \beta^t \mathbf{1}\{s_t \in \mathcal{F}_\pi\} \right] \\ \text{subject to} \quad & \pi_r(s) = \pi(s), \quad \forall s \in \mathcal{F}_\pi \\ & a_t = \pi_r(s_t), \quad \forall t = 0, 1, \ldots, \infty \end{aligned} \tag{7}$$

The key to solving this problem is to break the coupling between the policy and the recoverable set, simultaneously estimating $\mathcal{R}_\pi$ and obtaining an edited policy $\pi_r^*$ that achieves the fastest recovery. Using equations (2), (4) and (5), we can obtain the optimal solution to equation (7) (see Theorem 1 and its proof), ensuring closed-loop safety under recoverable disturbances while leaving the original policy's outputs unchanged within the feasible set.

The following sections describe the joint training of the recoverable set and edited policy in equation (4), their optimality with respect to equation (7), and the online application of the edited policy in equation (5).

## Recoverable set estimation and policy editing

This section first introduces the definition and estimation of the original policy's feasible set. On this basis, we define the policy's recoverable set and design a gated residual policy structure. Finally, we describe the joint solution of the recoverable set and edited policy, enabling the edited policy to achieve the fastest recovery within the maximal recoverable set and attain optimality in equation (7).

**Estimating the original policy's feasible set.** The feasible set $\mathcal{F}_\pi$ of the original policy $\pi$ is specified by the feasibility indicator function $F_\pi : \mathcal{S} \to \{0, 1\}$, defined as

$$F_\pi(s) = \mathbf{1}\{s_t \in \mathcal{S}_{\text{safe}}, a_t = \pi(s_t), t = 0, 1, \ldots, \infty, s_0 = s\} \tag{8}$$

The indicator equals 1 for states that can remain safe indefinitely under policy $\pi$ and 0 otherwise. The feasible set can accordingly be expressed as

$$\mathcal{F}_\pi = \{s \in \mathcal{S} \mid F_\pi(s) = 1\} \tag{9}$$

To estimate $F_\pi$, we first initialize an indicator function that equals 1 throughout the safe state set:

$$F^0(s) = \mathbf{1}\{s \in \mathcal{S}_{\text{safe}}\} \tag{10}$$

This initial function indicates only the safety of the current state and cannot reflect closed-loop safety under policy $\pi$. We therefore update it using the Bellman operator $\mathcal{T}_\pi$ in equation (2):

$$F^{k+1}(s) = (\mathcal{T}_\pi F^k)(s), \quad \forall s \in \mathcal{S} \tag{11}$$

As $k \to \infty$, this iteration converges pointwise to the feasibility indicator $F_\pi$ defined in equation (8) (see the proof of Theorem 2).

**Policy recoverable set and gated residual policy design.** Unlike existing methods that fine-tune a given policy for safety over the full state space or in hazardous states, we introduce the concept of a recoverable set to enable precise safety repair of the original policy. Outside the original policy's feasible set $\mathcal{F}_\pi$, there are three types of states: states that are already unsafe, states within the safe set that cannot return to $\mathcal{F}_\pi$, and states that can return to $\mathcal{F}_\pi$ through a finite sequence of recovery actions. The first two types cannot obtain closed-loop safety guarantees by returning to the original policy's feasible set, and optimizing them introduces unattainable objectives into training. The full set of states of the third type is the recoverable set of $\pi$, defined in equation (1). This set is policy-relative: the recovery target is not an arbitrary safe state, but $\mathcal{F}_\pi$, where the original policy $\pi$ already provides closed-loop safety. Subsequent fine-tuning therefore needs only to learn how to bring the vehicle back within the original policy's range of capabilities, rather than re-learn a complete driving policy.

To strictly satisfy the constraints in equation (7), we design the edited policy $\pi_r$ as a combination of the original policy $\pi$ and a state-gated residual branch,

$$\Pi = \left\{ \pi_r \;\middle|\; \begin{array}{ll} \pi_r(s) = \pi(s) + g(s)\delta(s), & \forall s \in \mathcal{S}, \\ g(s) = 0, & \forall s \in \mathcal{F}_\pi \end{array} \right\} \tag{12}$$

where $\pi$ is the frozen original policy and $\delta$ is the residual recovery action to be learned. The gating function $g(s) : \mathcal{S} \to \{0, 1\}$ is constrained to remain closed within the feasible set, leaving the original policy's outputs unchanged in this region. Outside $\mathcal{F}_\pi$, the gating function requires different designs for policy training and deployment. Because recoverable set estimation depends on optimization of the edited policy, the residual branch is enabled for all safe states $\mathcal{S}_{\text{safe}} \setminus \mathcal{F}_\pi$ during training, allowing optimization to discover recoverable states that have not yet been identified. During deployment, gate activation additionally requires the state to be judged recoverable, as detailed in the section on online application of the edited policy.

**Recoverable set estimation and edited policy optimization.** For any given trajectory $(s_0, a_0, s_1, \ldots)$, the first return time to the given policy's feasible set, $\tau_{\mathcal{F}_\pi}$, and the first entry time into the unsafe set, $\tau_{\mathcal{S}_{\text{unsafe}}}$, are defined as

$$\tau_{\mathcal{F}_\pi} = \inf\{t \geq 0 : s_t \in \mathcal{F}_\pi\}, \quad \tau_{\mathcal{S}_{\text{unsafe}}} = \inf\{t \geq 0 : s_t \in \mathcal{S}_{\text{unsafe}}\} \tag{13}$$

For any candidate edited policy $\pi_r$, define its recoverable set with respect to the given policy $\pi$ as

$$\mathcal{R}_\pi^{\pi_r} = \{s \in \mathcal{S} \setminus \mathcal{F}_\pi : \tau_{\mathcal{F}_\pi} < \tau_{\mathcal{S}_{\text{unsafe}}}, \; \tau_{\mathcal{F}_\pi} < \infty, s_0 = s, a_t = \pi_r(s_t), t = 0, \ldots, \infty\} \tag{14}$$

By the definition in equation (1), the recoverable set $\mathcal{R}_\pi$ can be written as

$$\mathcal{R}_\pi = \bigcup_{\pi_r \in \Pi} \mathcal{R}_\pi^{\pi_r} \tag{15}$$

Hereafter, we refer to $\mathcal{R}_\pi$ as the maximal recoverable set.

Our goal is to find an optimal edited policy $\pi_r^*$ whose induced recoverable set with respect to policy $\pi$ satisfies $\mathcal{R}_\pi^* = \mathcal{R}_\pi$. Analogously to estimating policy $\pi$'s feasible set, we estimate the recoverable set by defining the recoverability indicator function of policy $\pi_r$ with respect to policy $\pi$ (equation (3)), which has the following properties:

$$\begin{aligned} \mathcal{R}_\pi^{\pi_r} &= \{s \in \mathcal{S} \setminus \mathcal{F}_\pi : G_\pi^{\pi_r}(s) > 0\} \\ \mathcal{F}_\pi &= \{s \in \mathcal{S} : G_\pi^{\pi_r}(s) = 1\} \end{aligned} \tag{16}$$

We construct the optimal recoverability indicator function as

$$G_\pi^*(s) = \sup_{\pi_r \in \Pi} G_\pi^{\pi_r}(s), \quad \forall s \in \mathcal{S} \tag{17}$$

The set corresponding to this indicator function can be shown to be the maximal recoverable set $\mathcal{R}_\pi$. Furthermore, there exists an edited policy whose recoverability indicator function equals $G_\pi^*$, and any such policy achieves the fastest recovery (Theorem 4):

$$\mathcal{R}_\pi = \{s \in \mathcal{S} \setminus \mathcal{F}_\pi : G_\pi^*(s) > 0\} \tag{18}$$

To estimate $G_\pi^*$, we construct the recovery Bellman operator $\mathcal{T}_{\pi,\text{rec}}$ (equation (4)). This operator can be shown to be a contraction mapping that converges to the unique fixed point $G_\pi^*$ from any initial recoverability indicator function $G$ (see Theorem 1 for the contraction and convergence results). The max operation implicitly performs statewise policy optimization. To solve explicitly for the edited policy, we use policy iteration to decompose this operator into alternating policy evaluation and policy improvement:

Given the edited policy $\pi_r^k \in \Pi$ at outer iteration $k$, we fix its action selection and define the policy evaluation operator

$$(\mathcal{T}_{\pi,\text{PEV}}^k G)(s) = \begin{cases} 1, & s \in \mathcal{F}_\pi, \\ 0, & s \in \mathcal{S}_{\text{unsafe}}, \\ \beta G(f(s, \pi_r^k(s))), & \text{otherwise} \end{cases} \tag{19}$$

To distinguish outer policy iteration from inner policy evaluation, we start from any initial function $G^{k,0} : \mathcal{S} \to [0,1]$ and perform

$$G^{k,\ell+1}(s) = (\mathcal{T}_{\pi,\text{PEV}}^k G^{k,\ell})(s), \qquad \ell = 0, 1, \dots \tag{20}$$

The inner policy evaluation operator can be shown to converge to its unique fixed point $G_\pi^k := G_\pi^{\pi_r^k}$.

After completing policy evaluation at iteration $k$, we perform greedy improvement only at states where recovery has not yet terminated, obtaining the policy at iteration $k+1$:

$$\pi_r^{k+1}(s) = \begin{cases} \pi_r^k(s), & s \in \mathcal{F}_\pi \cup \mathcal{S}_{\text{unsafe}} \\ \arg\max_{a \in \mathcal{A}} G_\pi^k(f(s,a)), & \text{otherwise} \end{cases} \tag{21}$$

Here, $\mathcal{A}$ denotes the action space. For $s \in \mathcal{F}_\pi$, the policy outputs remain unchanged to satisfy the capability-preservation constraint in equation (12). The set $\mathcal{S}_{\text{unsafe}}$ is a terminal boundary, and actions taken there do not affect the recoverability function.

We alternate policy evaluation in equation (20) and policy improvement in equation (21). The recoverability indicator function converges uniformly:

$$\lim_{k \to \infty} \|G_\pi^k - G_\pi^*\|_\infty = 0 \tag{22}$$

Theorem 1 establishes this convergence and shows that the resulting policy is an optimal edited policy $\pi_r^*$.

## Neural network implementation and training

The Bellman relations and policy iteration above provide the theoretical approach to feasible set estimation and policy editing. This section describes the neural network implementation and training procedure.

**Feasibility indicator network for the original policy.** The Bellman iteration in equation (11) excludes states that lead to safety failures under the original policy by propagating failure information backwards. To approximate the resulting feasibility indicator in high-dimensional state spaces, we train a neural network using labels derived from finite closed-loop rollouts. If a trajectory encounters no safety failures (collisions or departures from the drivable area) over its full time horizon, all closed-loop states along it are labelled 1. If a safety failure occurs, all states from the start to the first failure are labelled 0. The feasibility indicator network $F$ takes the original model's feature encoding as input and outputs an estimate of the feasibility indicator function. It is trained using binary cross-entropy loss to approximate the original policy's closed-loop safety from each state. These labels reflect safety over the observed rollout horizon and provide a finite-horizon approximation to $F_\pi$.

**Recoverability indicator network.** The recoverability network $G$ is initialized from the weights of $F$ and serves as the value network during training of the edited policy. $G$ assesses the edited policy's ability to return safely to the original policy's feasible set from states outside it. Sampling therefore focuses on recovery trajectories from states outside the feasible set. When the closed-loop state is judged to have left the feasible set ($F < 0.3$), the gate opens and the edited policy performs safe recovery through residual corrections. The episode terminates when the state enters the unsafe region or is judged to have returned to the original policy's feasible set ($F > 0.9$). Different thresholds are used to identify departure from and return to the feasible set, improving the precision of these decisions. The collected samples are used to construct a TD-lambda optimization target for the recoverability indicator network, improving training efficiency and stability:

$$G_{\text{target}}(s_t) = r_t + \gamma(1-d_t)[(1-\lambda)G(s_{t+1}) + \lambda G_{\text{target}}(s_{t+1})], \qquad G_{\text{target}}(s_T) = 0,$$
$$L_G = \mathbb{E}_{s_t}\left[\left(G(s_t) - G_{\text{target}}(s_t)\right)^2\right] \tag{23}$$

Here, $r_t$ denotes a sparse reward for transition $(s_t, a_t, s_{t+1})$, taking the value 1 upon successful recovery to the original policy's feasible set and 0 otherwise. The flag $d_t \in \{0,1\}$ marks the end of a recovery training segment. The discount factor is $\gamma = 0.99$, and $\lambda = 0.95$ controls the weighting of multistep returns in the TD-lambda target. The time index within a contiguous segment is $t = 0, \ldots, T-1$, with $d_{T-1} = 1$. $G_{\text{target}}$ is the TD-lambda training target. $L_G$ is the loss function for $G$.

**Optimization of the edited policy.** We optimize the edited policy using the same sampled data as for training the recoverability indicator network. The gate remains open on these samples, and the policy corrects the original policy's outputs through the residual branch to maximize the recoverability indicator function. We update the residual branch using PPO's clipped surrogate objective[45] to ensure stable neural network training. Throughout optimization, the original policy's parameters remain frozen, and only the parameters of the residual branch are updated.

$$L_{\pi_r} = -\mathbb{E}_{s_t}\left[\min\left\{\frac{\pi_r(a_t \mid s_t)}{\pi_{r,\text{old}}(a_t \mid s_t)}A_t, \text{clip}\left(\frac{\pi_r(a_t \mid s_t)}{\pi_{r,\text{old}}(a_t \mid s_t)}, 1-\epsilon, 1+\epsilon\right)A_t\right\}\right] \tag{24}$$

Here, $\pi_{r,\text{old}}$ is the sampling policy, $A_t = G_{\text{target}}(s_t) - G(s_t)$ is the advantage estimate, $\epsilon = 0.2$ is the clipping coefficient, and $L_{\pi_r}$ is the loss function of the edited policy.

## Online application of the edited policy

This section introduces the online application of RPE, which reuses the given policy and guarantees closed-loop safety under recoverable disturbances. The main idea is to use the feasibility and recoverability indicator functions to control switching of the policy gate between open and closed states.

The gate follows the switching logic in equation (5). The edited policy operates online as follows. When the autonomous vehicle lies within the original policy's feasible set, the gate is closed and the vehicle uses the original policy. When a state transition inconsistent with the training environment causes the vehicle to leave the feasible set, $G^*_\pi(s) > 0$ indicates that the current state is recoverable. The gate then opens immediately, and the edited policy uses residual corrections to follow the fastest safe recovery trajectory. If another disturbance occurs before recovery is complete, the gate continues to operate according to the resulting state. During recovery, the gate remains open until the state returns to the feasible set. It then closes, and the original policy resumes. Theorem 3 establishes closed-loop safety provided that every affected state transition leaves the vehicle in $\mathcal{F}_\pi \cup \mathcal{R}_\pi$.

The theoretical gate above uses the feasibility indicator function $F_\pi$ and the optimal recoverability indicator function $G^*_\pi$. In practical online operation, their neural network estimates $F$ and $G$ control the gate instead: the gate opens when the vehicle state is judged to have left the original policy's feasible set and to remain recoverable ($F < 0.3$ and $G \geq 0.1$), and closes when the vehicle is judged to have returned to the feasible set ( $F > 0.9$). The decision thresholds for $F$ remain the same as in training of the edited policy, reducing false positives in gate opening and closing. The only additional entry condition is $G \geq 0.1$, which restricts the scope of intervention by the residual branch.

## Theorem 1

This theorem establishes the main result of this work: our method yields an optimal edited policy $\pi_r^*$ that achieves the fastest safe recovery to the original policy's feasible set from every state in the maximal recoverable set $\mathcal{R}_\pi$.

**Proof of Theorem 1.** We first prove that iteration of the recovery Bellman operator converges to $G_\pi^*$, enabling identification of all recoverable states. For any bounded functions $G$ and $\widetilde{G}$, if $s \in \mathcal{F}_\pi \cup \mathcal{S}_{\text{unsafe}}$, then $\mathcal{T}_{\pi,\text{rec}} G(s) = \mathcal{T}_{\pi,\text{rec}} \widetilde{G}(s)$. For the remaining states,

$$\begin{aligned} &\left|(\mathcal{T}_{\pi,\text{rec}} G)(s) - (\mathcal{T}_{\pi,\text{rec}} \widetilde{G})(s)\right| \\ &\quad \le \beta \max_{a \in \mathcal{A}} \left| G(f(s,a)) - \widetilde{G}(f(s,a)) \right| \\ &\quad \le \beta \|G - \widetilde{G}\|_\infty \end{aligned} \tag{25}$$

Thus, $\|\mathcal{T}_{\pi,\text{rec}} G - \mathcal{T}_{\pi,\text{rec}} \widetilde{G}\|_\infty \le \beta \|G - \widetilde{G}\|_\infty$. By the Banach fixed-point theorem, $\mathcal{T}_{\pi,\text{rec}}$ has a unique fixed point, and the corresponding fixed-point iteration converges from any bounded initial function. By the Bellman optimality relation in equations (17) and (4), this fixed point is $G_\pi^*$.

We next prove that alternating policy evaluation and policy improvement yields the optimal recoverability indicator function and the optimal edited policy. For a fixed edited policy $\pi_r^k$, equation (19) replaces the maximization operation with $\pi_r^k(s)$. Similarly, we have

$$\left\| \mathcal{T}_{\pi,\text{PEV}}^k G - \mathcal{T}_{\pi,\text{PEV}}^k \widetilde{G} \right\|_\infty \le \beta \left\| G - \widetilde{G} \right\|_\infty \tag{26}$$

Hence, the current edited policy's recoverability function $G_\pi^{\pi_r^k}$ is the unique fixed point of this operator, and the inner iteration in equation (20) converges to it.

We further prove that each improved policy $\pi_r^{k+1}$ obtained from equation (21) has an improved recoverability indicator function. By the improved policy's greedy action selection,

$$(\mathcal{T}_{\pi,\text{PEV}}^{k+1} G_\pi^k)(s) = (\mathcal{T}_{\pi,\text{rec}} G_\pi^k)(s) \ge (\mathcal{T}_{\pi,\text{PEV}}^k G_\pi^k)(s) = G_\pi^k(s), \quad \forall s \in \mathcal{S} \tag{27}$$

To show that the one-step improvement in equation (27) propagates through the full policy evaluation of $\pi_r^{k+1}$, we prove that the function sequence $G^{k+1,\ell}$ increases monotonically during policy evaluation initialized at $G^{k+1,0} = G_\pi^k$. Equation (27) first gives $G^{k+1,1}(s) \ge G_\pi^k(s) = G^{k+1,0}(s), \forall s \in \mathcal{S}$. If $G^{k+1,\ell+1} \ge G^{k+1,\ell}$, then for $s \notin \mathcal{F}_\pi \cup \mathcal{S}_{\text{unsafe}}$,

$$\begin{aligned} &G^{k+1,\ell+2}(s) - G^{k+1,\ell+1}(s) \\ &\quad = \left(\mathcal{T}_{\pi,\text{PEV}}^{k+1} G^{k+1,\ell+1}\right)(s) - \left(\mathcal{T}_{\pi,\text{PEV}}^{k+1} G^{k+1,\ell}\right)(s) \\ &\quad = \beta \left(G^{k+1,\ell+1}(f(s,\pi_r^{k+1}(s))) - G^{k+1,\ell}(f(s,\pi_r^{k+1}(s)))\right) \ge 0 \end{aligned} \tag{28}$$

By induction, the sequence increases monotonically and converges to $G_\pi^{k+1}(s)$. Therefore,

$$G_\pi^{k+1}(s) \ge G_\pi^k(s), \quad \forall s \in \mathcal{S} \tag{29}$$

The same monotonicity argument gives $G_\pi^{k+1} \ge \mathcal{T}_{\pi,\text{rec}} G_\pi^k$. Since $G_\pi^{k+1} \le G_\pi^*$ and $\mathcal{T}_{\pi,\text{rec}} G_\pi^* = G_\pi^*$, contraction yields

$$\|G_\pi^* - G_\pi^{k+1}\|_\infty \le \|\mathcal{T}_{\pi,\text{rec}} G_\pi^* - \mathcal{T}_{\pi,\text{rec}} G_\pi^k\|_\infty \le \beta \|G_\pi^* - G_\pi^k\|_\infty \tag{30}$$

Thus, $G_\pi^k$ converges uniformly to $G_\pi^*$. A final policy improvement step with respect to this limiting function yields an edited policy $\bar{\pi}_r$. By its greedy action selection and the Bellman fixed-point relation,

$$\mathcal{T}_{\pi,\text{PEV}}^{\bar{\pi}_r} G_\pi^* = \mathcal{T}_{\pi,\text{rec}} G_\pi^* = G_\pi^* \tag{31}$$

Since the policy evaluation operator has a unique fixed point, $G_\pi^{\bar{\pi}_r} = G_\pi^*$. Thus, $\bar{\pi}_r$ induces the optimal recoverability indicator function. By Theorem 4, it is an optimal edited policy $\pi_r^*$ and achieves the fastest safe recovery throughout the maximal recoverable set. This completes the proof.

### Theorem 2

This theorem establishes the convergence and optimality of the policy feasibility indicator function $F_\pi$, providing the basis for recoverable set estimation and policy editing.

**Proof of Theorem 2.** Recall that $f_\pi(s) = f(s, \pi(s))$ is the deterministic closed-loop next state induced by the given policy $\pi$. Let

$$\mathcal{F}^k = \{s \in \mathcal{S} : F^k(s) = 1\} \tag{32}$$

denote the set of states still classified as feasible after iteration $k$. We first show that $\mathcal{F}^k$ has the following finite-horizon interpretation:

$$\mathcal{F}^k = \{s \in \mathcal{S} : s_t \in \mathcal{S}_{\text{safe}},\ t = 0, 1, \ldots, k,\ s_{t+1} = f_\pi(s_t),\ s_0 = s\} \tag{33}$$

For $k = 0$, equation (10) gives $\mathcal{F}^0 = \mathcal{S}_{\text{safe}}$, so the claim holds. Suppose that it holds for $k$. Equations (2) and (11) imply that $F^{k+1}(s) = 1$ if and only if $s \in \mathcal{S}_{\text{safe}}$ and $F^k(f_\pi(s)) = 1$. By the induction hypothesis, the latter condition is equivalent to all states over the first $k$ steps from $f_\pi(s)$ remaining safe, or equivalently, all closed-loop states over the first $k+1$ steps from $s$ remaining safe. Thus, the expression holds for every $k$.

The finite-horizon interpretation above implies $\mathcal{F}^{k+1} \subseteq \mathcal{F}^k$, or equivalently, $F^{k+1}(s) \leq F^k(s)$ for every $s \in \mathcal{S}$. Thus, $\{F^k(s)\}_{k=0}^\infty$ is a monotone sequence taking values in $\{0, 1\}$, and the pointwise limit

$$F^\infty(s) = \lim_{k \to \infty} F^k(s) \tag{34}$$

must exist. In a finite-state or discretized implementation, this monotone elimination process converges in finitely many iterations. In a general state space, the limit above establishes pointwise convergence.

We next prove that the limiting function equals the closed-loop safety indicator defined in equation (8). By the definition of $\mathcal{F}^k$, $F^\infty(s) = 1$ if and only if $s$ belongs to every finite-horizon safe set $\mathcal{F}^k$. In other words, the closed-loop trajectory induced by policy $\pi$ from $s$ never leaves $\mathcal{S}_{\text{safe}}$ over any finite horizon. This is equivalent to

$$s_t \in \mathcal{S}_{\text{safe}}, \quad s_{t+1} = f_\pi(s_t), \quad t = 0, 1, \ldots, \quad s_0 = s \tag{35}$$

that is, $F_\pi(s) = 1$. Conversely, if $F_\pi(s) = 1$, the trajectory is safe over every finite horizon, so $s \in \mathcal{F}^k, \forall k$, and hence $F^\infty(s) = 1$. Therefore, $F^\infty = F_\pi$. This completes the proof.

### Theorem 3

Theorem 3 proves that the online application of the edited policy in equation (5) provides closed-loop safety under any recoverable disturbance.

**Proof of Theorem 3.** Let $t_d$, $t_g$ and $t_r$ denote the time of a disturbance, the first gate opening and the completion of recovery, respectively. Before the disturbance, the vehicle lies in $\mathcal{F}_\pi$ and executes policy $\pi$, so it remains safe. For any post-disturbance state $s_{t_d} \in \mathcal{R}_\pi$, the definition of the maximal recoverable set gives $F_\pi(s_{t_d}) = 0$ and $G_\pi^*(s_{t_d}) > 0$. According to equation (5), the gate opens immediately, so $t_g = t_d$. The edited policy $\pi_r^*$ then performs the fastest recovery through residual corrections, with the gate remaining open until the state returns to $\mathcal{F}_\pi$. By $G_\pi^{\pi_r^*} = G_\pi^*$ and the definition of recoverability, there exists a finite time $t_r > t_g$ at which the vehicle returns to $\mathcal{F}_\pi$ before entering the unsafe set. The recovery phase is therefore safe, with all intermediate states remaining in $\mathcal{R}_\pi$. Upon reaching $\mathcal{F}_\pi$, the gate closes and execution of $\pi$ resumes. By the definition of $\mathcal{F}_\pi$, executing $\pi$ preserves this set in the absence of further disturbances. Thus, the online gate preserves $\mathcal{F}_\pi \cup \mathcal{R}_\pi \subseteq \mathcal{S}_{\text{safe}}$ and ensures closed-loop safety, provided that disturbances leave the state in this union. This completes the proof.

### Theorem 4

This theorem proves that the optimal recoverability indicator function $G_\pi^*$ characterizes the maximal recoverable set $\mathcal{R}_\pi$ and that there exists an edited policy $\pi_r^*$ that induces this indicator function and achieves the fastest recovery. Furthermore, any policy that induces this indicator function achieves the fastest recovery.

**Proof of Theorem 4.** We first show that $G_\pi^*$ characterizes the maximal recoverable set. By the definitions of $G_\pi^*$ and $\mathcal{R}_\pi$, it suffices to prove

$$\{s \in \mathcal{S} \setminus \mathcal{F}_\pi : \sup_{\pi_r \in \Pi} G_\pi^{\pi_r}(s) > 0\} = \bigcup_{\pi_r \in \Pi} \{s \in \mathcal{S} \setminus \mathcal{F}_\pi : G_\pi^{\pi_r}(s) > 0\} \tag{36}$$

First, if $s$ belongs to the union on the right-hand side, then there exists an edited policy $\pi_r$ such that $G_\pi^{\pi_r}(s) > 0$. Thus,

$$\sup_{\pi_r \in \Pi} G_\pi^{\pi_r}(s) \geq G_\pi^{\pi_r}(s) > 0 \tag{37}$$

so $s$ belongs to the set on the left-hand side.

Conversely, if $s$ belongs to the set on the left-hand side, then

$$\sup_{\pi_r \in \Pi} G_\pi^{\pi_r}(s) > 0 \tag{38}$$

If no edited policy satisfies $G_\pi^{\pi_r}(s) > 0$, then, because all $G_\pi^{\pi_r}(s) \geq 0$, we must have $G_\pi^{\pi_r}(s) = 0, \forall \pi_r \in \Pi$. Their supremum would therefore also be 0, contradicting the expression above. Hence, there must exist an edited policy $\pi_r$ such that $G_\pi^{\pi_r}(s) > 0$, meaning that $s$ belongs to the union on the right-hand side. The two sets are therefore equal.

We next construct an edited policy in $\Pi$. For each state $s$, we can select a maximizing policy,

$$\bar{\pi}_s \in \arg\max_{\pi_r \in \Pi} G_\pi^{\pi_r}(s) \tag{39}$$

choosing any maximizer if it is not unique. We then construct the edited policy state by state as

$$\pi_r^*(s) = \bar{\pi}_s(s), \quad \forall s \in \mathcal{S} \tag{40}$$

Because $\bar{\pi}_s \in \Pi$ for every $s \in \mathcal{S}$, $\pi_r^*$ also has the gated residual form $\pi_r^*(s) = \pi(s) + g^*(s)\delta^*(s)$, with $g^*(s) = 0$ when $s \in \mathcal{F}_\pi$. Thus, $\pi_r^* \in \Pi$, establishing existence.

We now prove that this policy induces $G_\pi^*$ and achieves the fastest recovery. For any policy $\pi_r \in \Pi$, define the time to return to $\mathcal{F}_\pi$ from state $s$ before entering the unsafe set as

$$T_{\pi_r}(s) = \begin{cases} \tau_{\mathcal{F}_\pi}, & \tau_{\mathcal{F}_\pi} < \tau_{\mathcal{S}_{\text{unsafe}}}, \tau_{\mathcal{F}_\pi} < \infty, \\ \infty, & \text{otherwise} \end{cases} \tag{41}$$

The recoverability indicator function of $\pi_r$ with respect to $\pi$ can then be written as

$$G_\pi^{\pi_r}(s) = \beta^{T_{\pi_r}(s)} \tag{42}$$

Thus, maximizing $G_\pi^{\pi_r}(s)$ is equivalent to minimizing the return time $T_{\pi_r}(s)$. Define the shortest safe return time for every state as

$$T^*(s) = \min_{\pi_r \in \Pi} T_{\pi_r}(s) \tag{43}$$

Thus, $T^*(s) = 0$ on $\mathcal{F}_\pi$, $T^*(s) < \infty$ on $\mathcal{R}_\pi$, and $T^*(s) = \infty$ elsewhere. By the definition of $\bar{\pi}_s$, using $\bar{\pi}_s$ from $s \in \mathcal{R}_\pi$ achieves the fastest return, and its first action $\pi_r^*(s)$ lies on a shortest recovery trajectory. If $T^*(s) = 1$, recovery is complete after executing $\pi_r^*(s)$. If $T^*(s) = k > 1$, a contradiction argument shows that the next state $s' = f(s, \pi_r^*(s))$ satisfies $T^*(s') = k - 1$. Otherwise, a shorter safe continuation from $s'$ would yield a shorter recovery path from $s$, contradicting the minimality of $T^*(s)$. By induction, $\pi_r^*$ returns to $\mathcal{F}_\pi$ in $T^*(s)$ steps from any $s \in \mathcal{R}_\pi$, proving that it achieves the fastest recovery.

We next show that policy $\pi_r^*$ induces $G_\pi^*$. For every $s \in \mathcal{R}_\pi$,

$$G_\pi^{\pi_r^*}(s) = \beta^{T^*(s)} = G_\pi^*(s) \tag{44}$$

On the other hand, if $s \notin \mathcal{R}_\pi \cup \mathcal{F}_\pi$, the definition of the maximal recoverable set implies that no candidate edited policy can return $s$ safely to $\mathcal{F}_\pi$ in finite time. Hence, $G_\pi^*(s) = 0$ and $G_\pi^{\pi_r^*}(s) = 0$. If $s \in \mathcal{F}_\pi$, both equal 1. Therefore,

$$G_\pi^{\pi_r^*}(s) = G_\pi^*(s), \quad \forall s \in \mathcal{S} \tag{45}$$

Finally, we prove that any policy $\mu$ inducing $G_\pi^*$ achieves the fastest recovery:

$$G_\pi^\mu(s) = G_\pi^*(s) = \beta^{T^*(s)} = \max_{\mu \in \Pi} \beta^{T_\mu(s)}, \quad \forall s \in \mathcal{S} \tag{46}$$

This completes the proof.